\documentclass[letterpaper, 10pt, conference]{ieeeconf}  
\usepackage[normalem]{ulem}
\usepackage{url}

\IEEEoverridecommandlockouts                              

\newif\iftrversion\trversiontrue

\newcommand{\hadi}[1]{{\color{red}}{\color[rgb]{0,0,0}#1}}

\usepackage{macros}

\newtheorem{theo}{Theorem}
\newtheorem{theorem}[theo]{Theorem}

\newtheorem{definition}{Definition}
\newcommand\mps{\textsf{m/s}}

\title{\LARGE \bf
Path-Following through Control Funnel Functions
}

\author{Hadi Ravanbakhsh, Sina Aghli, Christoffer Heckman, and Sriram Sankaranarayanan$^\ast$
\thanks{This work was supported by NSF award no.\ 1646556.}%
\thanks{All authors are with the Department of Computer Science, University of Colorado, Boulder, CO 80309 USA}%
\thanks{*Corresponding author. E-mail address: {\tt\small srirams at colorado.edu}}
}

\begin{document}

\maketitle
\thispagestyle{empty}
\pagestyle{empty}

\begin{abstract}
  We present an approach to path following using so-called control
  funnel functions. Synthesizing controllers to ``robustly'' follow a reference
  trajectory is a fundamental problem for autonomous vehicles. Robustness, in this context,
  requires our controllers to handle a specified amount of deviation from
  the desired trajectory. Our approach  considers a timing law that
  describes how fast to move along a given reference trajectory and a
  control feedback law for reducing deviations from the reference. We synthesize
  both feedback laws using ``control funnel functions'' that jointly encode the
  control law as well as its correctness argument over a mathematical
  model of the vehicle dynamics. We adapt a previously described
  demonstration-based learning algorithm to synthesize a control
  funnel function as well as the associated feedback law. We implement
  this law on top of a 1/8th scale autonomous vehicle called the
  Parkour car. We compare the performance of our path following
  approach against a trajectory tracking approach by specifying
  trajectories of varying lengths and curvatures. Our experiments
  demonstrate the improved robustness obtained from the use of
  control funnel functions.
\end{abstract}

\section{INTRODUCTION}~\label{sec:intro}
Recent advances in motion planning have brought truly autonomous systems closer to
becoming a reality. However, one of the main challenges is their
safety. Plan execution may fail because of considerable uncertainties
such as disturbances, imprecise measurements, and modeling. Such
failures can lead to safety violation with catastrophic
consequences. Given a reference trajectory with associated timing, a
straightforward solution is to design a feedback control which tracks
the reference trajectory and reduces tracking errors. Another idea is
to use ``path-following" where the goal is to track a path without
timing constraints. Path-following techniques provide smooth
convergence to the reference trajectory \hadi{while avoiding input
saturation~\cite{HAUSER1995,Pappas1996}, and are more robust with 
respect to measurement errors and external disturbances~\cite{Egerstedt2001virtual}}. In this paper, we investigate
path-following techniques to improve robustness and safety for plan
execution \hadi{through control funnel functions.}

\begin{figure}[t]
\begin{center}
	\includegraphics[width=0.4\textwidth]{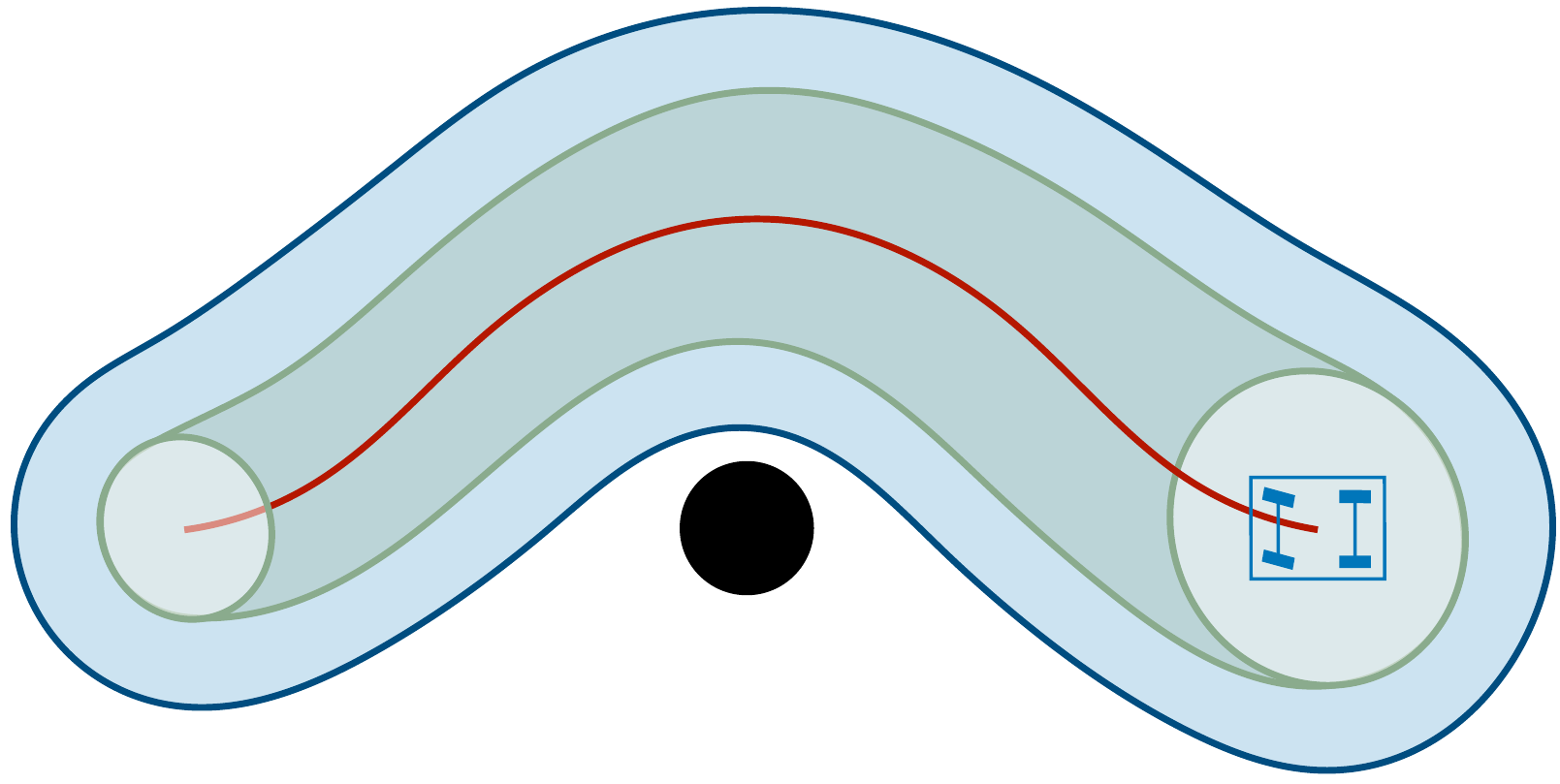}
\end{center}
\caption{Safe tracking with control funnels: the black box depicts an
  obstacle, solid red line shows the reference trajectory generated by
  the planner. The projection of the funnel on $x$-$y$ plane is shown
  in green.  The green circle at the right is the head of the funnel
  and the one to the left is the tail. The blue region enlarges the
  funnel to account for the non-zero size of the
  vehicle.}\label{fig:safety-funnel}
\end{figure}

First, we reformulate the problem as a path following problem that
adds an extra control input to the system to decide how fast to move
along the reference trajectory. This naturally allows our approach to
speed up/slow down progress along the reference trajectory.
Furthermore, our method is based on 
control funnel functions inspired
by the concepts of control
funnels~\cite{mason1985mechanics,majumdar2013control} and control
barrier functions~\cite{WIELAND2007462}. Given a reference trajectory
segment and a desired ``safe region'' around the trajectory, a control
funnel function guarantees that the system moves along one end of the
trajectory to another while remaining inside the safe region.  We use a
previously-developed framework to synthesize control funnels
automatically given the vehicle dynamics, the reference
trajectory, and the surrounding safe region~\cite{Ravanbakhsh-RSS-17}.

We have implemented our method using a standard single-track model
for capturing the ground vehicle dynamics.  The
resulting controllers are then tested on a $\frac{1}{8}^{th}$ scale
vehicle platform called the Parkour car developed at CU-Boulder.
Using our approach, we successfully synthesize controllers
for numerous reference trajectories and demonstrate the ability of the
controller synthesized on a toy model to drive an actual vehicle in
the lab. We also show that, when contrasted with trajectory tracking,
path following approaches provide a higher level of robustness as shown
by some of our experiments that involved large disturbances applied to
the car during motion.

\hadi{
In the rest of this section we review the literature. We discuss previously developed methods and our notations in Section~\ref{sec:background}. Sections~\ref{sec:main} and~\ref{sec:bounded} present our contributions. Finally, in Section~\ref{sec:expr}, we discuss our experimental results. 
}

\subsection{Related Work}
Stability for autonomous vehicles is a challenging problem. Brockett~\cite{brockett1983} showed that even a simple unicycle model cannot be stabilized using continuous feedback laws. However, continuous feedback laws exist for stabilization to non-stationary trajectories; these feedback laws are usually obtained through linearization~\cite{Walsh1994}. 
While trajectory tracking has been widely used to solve plan execution, it has several shortcomings which are addressed using path-following.
In pioneering work ~\cite{Nelson1990,Sampei1991,Canudas1991,Samson1992,Sordalen1993} the velocity of the vehicle tracked a desired reference velocity and the controller is designed to steer the vehicle to the path. These path-following methods have been shown to yield a smoother convergence to the trajectory while avoiding input saturation. Beside these works, a wide diversity of approaches are used to study the path-following problem. One line of work is based on designing vector fields surrounding the path to guarantee reaching and following the path~\cite{Nelson2007,lawrence2007}. Another approach is to use model predictive control~\cite{Faulwasser2009,Faulwasser2013}. In this article, we consider a line of effort distinct from these others.

Hauser et al.~\cite{HAUSER1995} proposed the conversion of the trajectory
tracking strategy to the so-called maneuver regulation strategy. The
main idea is to decrease the distance between the state and the
reference trajectory, not a specific point on the trajectory. The
reference trajectory $\vx_r(\cdot)$ is parameterized using a variable
$\theta$ (instead of time) and distance is defined as a function of
$\vx - \vx_r(\theta)$. $\theta$ and treated as a variable. An update
law (timing law) is then applied to ensure proper change of
$\theta$. Hauser and Hindman showed that this maneuver regulation
trick would yield a system that avoids input saturation. Similarly,
Pappas~\cite{Pappas1996} showed that by re-parameterizing the
trajectory, one could avoid input saturation. Subsequently, Encarnacao
et al.~\cite{Encarnacao2001} extended the technique for the output
maneuvering problem on a restricted set of dynamics. m

Following Hauser et al.~\cite{HAUSER1995}, others have divided the
task into two parts. The first task is to reach and follow the
reference trajectory using the variable $\theta$ (instead of time), and the
second task is to \textit{improve} the solution using an extra control input
$\theta$. For example, in Skjetne et al.~\cite{SKJETNE2004}, first the
system output is stabilized, and then a control law for $\theta$ is
used to adjust the velocity. In this work, we use the extra
freedom to control $\theta$ for increasing robustness. More
specifically, this extra degree of freedom allows us to design more
robust control Lyapunov functions (CLFs) from which we extract
the feedback as well as the timing law. 

Control Lyapunov functions were originally \hadi{introduced by
Sontag~\cite{sontag1989universal,sontag1983lyapunov}}. Synthesis
of CLFs is hard, involving bilinear matrix inequalities
(BMIs)~\cite{tan2004searching,majumdar2013control}.
Standard approaches such as alternating minimization result often do
not converge to a solution. To combat this, Majumdar et al.\ use
LQR controllers and their associated Lyapunov functions for the
linearization of the dynamics as good initial seed
solutions~\cite{majumdar2013control}. In contrast, recent work by some
of the authors remove the bilinearity by using a demonstrator in the form of a MPC
controller~\cite{Ravanbakhsh-RSS-17}. Furthermore, this approach
avoids local saddle points and has a fast convergence guarantee.

Aguiar et al.~\cite{AGUIAR2004} argue that there are performance
limitations for systems with unstable zero dynamics if one uses
trajectory tracking. However, using an extra control input
$\theta$, this restriction vanishes. The timing law in this work is
designed as a function of $\theta$ and its higher derivatives.

\hadi{Egerstedt et al.~\cite{Egerstedt2001virtual} develop a method where the reference point dynamics are governed by tracking error feedback.
Similarly}, Faulwasser et al.~\cite{Faulwasser2014} proposed designing the timing
law as a function of $\theta$, tracking error, and their higher derivatives. For example, one
can design a timing law which slows down the progress of $\theta$ when
the distance between the state and the reference is large.
They also combine the idea of path-following with control funnels. They  Similarly, we use control funnels to provide formal guarantees. However, the funnel is constructed using a CLF. Besides this, the timing law in our work is a function of the state $\vx$ and depends on the structure of the CLF.

\section{BACKGROUND}~\label{sec:background}
\begin{figure}[t]
\begin{center}
\vspace{0.2cm}
	\includegraphics[width=0.25\textwidth]{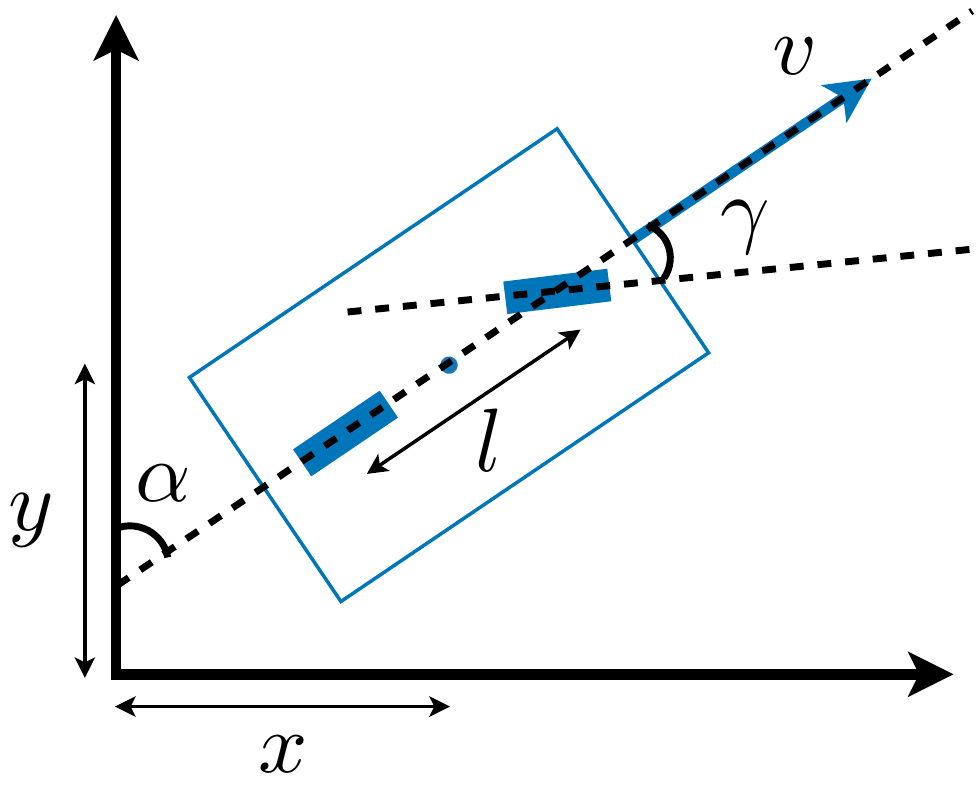}
\end{center}
\caption{A Schematic Diagram of the Bicycle Model.}\label{fig:bicycle-model} 
\end{figure}

This paper investigates CLF-based path following focusing on
applications to ground vehicles. We will use the well-known bicycle
model, whose state consists of its position ($x$ and $y$), its
orientation ($\alpha$), and velocity ($v$)\hadi{~\cite{FREGO201718,Egerstedt2001virtual}}. The rear axle is
perpendicular to the bicycle's axis, the front wheel's orientation can
be adjusted to steer the vehicle (see
Figure~\ref{fig:bicycle-model}). Let $\gamma$ be the angle between the
front axle and the bicycle axis (Fig.~\ref{fig:bicycle-model}). We
will assume that $\gamma \in [-\frac{\pi}{4}, \frac{\pi}{4}]$ is a
control input to the model. Also, the thrust applied to the vehicle is
$\T \in [-4, 4]$. The model has the following dynamics:
\begin{equation}\label{eq:bicycle-model}
  \begin{array}{l}
   \dot{x} \ =\ v \sin(-\alpha),\ \  \dot{y} \ =\ v \cos(\alpha) \\
    \dot{\alpha}\ =\ \frac{v}{l} \tan(\textcolor{blue}{\gamma}),\ \ \dot{v} \ =\  \textcolor{blue}{\T} \,,
    \end{array}
\end{equation}
wherein $l$ is the distance between the wheels and the control inputs
to the model $(\textcolor{blue}{\gamma}, \textcolor{blue}{\T})$ are shown in blue.

We will use this model to analyze the behavior of ground vehicles. In this
paper, we will study the design of control inputs to solve the problem
of controlling the vehicle to follow a given trajectory. Such trajectories are generated using planning algorithms such
as RRTs and are often designed to avoid obstacles in the
workspace~\cite{LaValle05}.

As an example, consider a scenario where the vehicle moving with speed
$2\mps$ needs to circumnavigate an obstacle as shown in
Fig.~\ref{fig:safety-funnel}. First, the planner generates a reference
trajectory shown with the solid red line. Note that, by design, the
reference trajectory keeps some distance from the obstacle.

To guarantee safety and trajectory tracking at the same time, we use a
control funnel~\cite{mason1985mechanics} (the green region in
Fig.~\ref{fig:safety-funnel}) which contains the reference
trajectory. The corresponding control law for a funnel guarantees that once
the state is inside the funnel, it remains in the funnel until it
reaches the desired target set of states. In other words, the system
safety in the tracking process is formally guaranteed.  In this work,
we wish to improve the process of funnel design to increase
robustness. Our funnel design technique is based on stability analysis
which is discussed first.

\subsection{Stability Analysis}
The dynamics for autonomous vehicles can be modeled with Euler
equations to study the behavior of these systems. In a continuous time
setting, the state of the system $\vx \in \reals^n$ updates w.r.t.\ an ordinary differential equation. Formally, $\dot{\vx} = f(\vx, \vu)$, 
where $\dot{\vx}$ is the derivative of $\vx$ w.r.t.\ time and $\vu \in \reals^m$ is the control input. 
Stability is a fundamental property of dynamical systems.  Numerous
control problems can be viewed as controlling a given system to
stabilize to a given equilibrium state $\vx_r$ or an ``equilibrium''
reference trajectory $\vx_r(t)$.  Control Lyapunov functions (CLF) are
a powerful tool for designing such stabilizing
controls~\cite{sontag1983lyapunov,artstein1983stabilization}. We first describe CLFs for
stabilizing to an equilibrium state.

\begin{definition}[Control Lyapunov Functions]\label{def:clf}
  A CLF $V$ is a smooth and radially unbounded function that maps each
  state to a real non-negative value, such that \textbf{(a)}
  $V(\vx_r) = 0$ and $V(\vx) > 0$ for all $\vx \not=\vx_r$, and
  \textbf{(b)}
  $(\forall\ \vx \not=\vx_r)\ (\exists \vu) \ \dot{V}(\vx, \vu) < 0$,
  wherein $\dot{V}$ is the Lie-derivative of $V$ w.r.t. $f$:
  $\dot{V}(\vx, \vu) =  \nabla V \cdot
  f(\vx, \vu)$.
\end{definition}
Condition (a) ensures that the value of $V$ is zero at
the equilibrium and strictly positive everywhere else. Condition (b)
ensures that for any state, we can find a control input that can achieve
an instantaneous decrease to the value of $V$. In this sense, CLFs are
equivalent to \emph{artificial potential functions} over the state
space~\cite{lopez1995autonomous}. Having a CLF, one can design a
feedback law which always decreases the value of $V$ and therefore
stabilizes the system to the equilibrium point. For instance, Sontag
provides a simple means to extract a feedback law from a given
CLF~\cite{sontag1989universal}.

\subsection{Trajectory Tracking vs.\ Path Following}

Another form of stability appears in trajectory tracking, wherein the
goal is to stabilize to a reference trajectory $\vx_r(t)$. Formally,
let $\vx_d(t):\ \vx(t) - \vx_r(t)$ describe the ``deviation'' from the
reference trajectory state at time $t$. The goal is to stabilize
$\vx_d$ to the equilibrium $\vzero$ under the time-varying reference frame
that places $\vx_r(t)$ as the origin at time $t$. The dynamics for
$\vx_d$ is defined as: $\dot{\vx}_d = f(\vx, \vu) - \vr(t)$, wherein
$\dot{\vx}_r = \frac{d\vx_r}{dt} = \vr(t)$.

One of the key drawbacks of trajectory tracking is that it specifies
the reference trajectory $\vx_r(t)$ along with the \emph{reference
  timing}, wherein the state $\vx_r(t)$ must ideally be achieved
at time $t$. This poses a challenge for control design unless the
timing is designed very carefully. Imagine, a reference trajectory
that traverses a winding hilly road at constant speeds. This compels
the control to constantly accelerate the vehicle on upslopes only
to ``slam the brakes'' on downhill sections~\cite{Hauser+Saccon/2006/Motorcycle}. 
 
Path following, on the other hand, separates these concerns by
allowing the user to specify a reference (feasible) path parameterized
with a scalar, $\theta$ (instead of time), $\vx_r(\theta)$ yields a
state for each $\theta$ and $\frac{d\vx_r}{d\theta} = \vr(\theta)$. 
As proposed by Hauser et al., one could define $\Pi$ as a function
that maps a state $\vx$ to the closest state on the reference
trajectory
$\vx_r(\cdot)$, using an auxiliary map $\pi$ ~\cite{HAUSER1995,Saccon+Hauser+Beghi/2013/Virtual}:
\[ \pi(\vx):\ \underset{\theta}{\mbox{argmin}} \ ||\vx -
  \vx_r(\theta)||_P^2 \ \,, \ \Pi(\vx) : \vx_r(\pi(\vx)) \] where $||\vx||_P^2 :\ \vx^t P \vx$ is a
Lyapunov function for the linearized dynamics around the reference
trajectory.  In order to stabilize the system to the reference path,
Hauser et al.\ propose to decrease the value of
$||\vx - \Pi(\vx)||_P^2$. However, as the projection function $\pi$
can get complicated, they use local approximations of $\pi$.

\begin{figure}[t]
\begin{center}
\vspace{0.2cm}
	\includegraphics[width=0.25\textwidth]{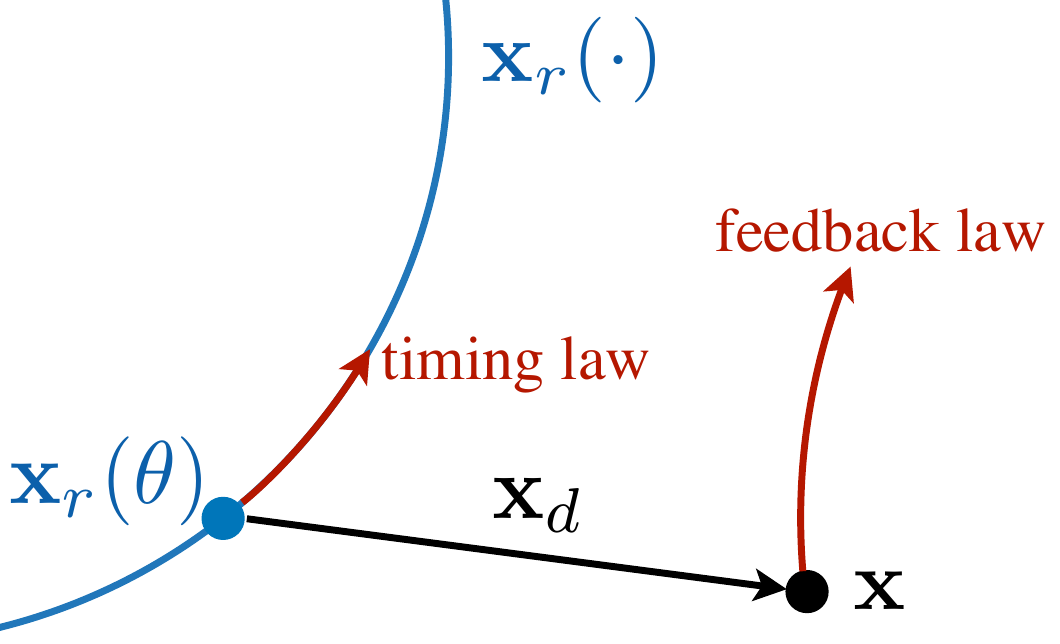}
\end{center}
\caption{Schematic View of a System along the Parameterized Path.}\label{fig:new-system} 
\end{figure}

Following this, others have proposed to design a control law for a
virtual input $u_0$ that controls $\theta$ as a function of time
(called the \emph{timing feedback law}), or in other words, the
progress (or sometimes regress) along the
reference~\cite{SKJETNE2004,AGUIAR2004}. Therefore, the deviation is
now defined as $\vx_d(t):\ \vx(t) - \vx_r(\theta(t))$ wherein
$\diff{\theta}{t} = u_0$. As depicted in Fig.~\ref{fig:new-system},
$\theta$ is mapped to a state on the path $\vx_r(\theta)$.
For example Faulwasser et al.~\cite{Faulwasser2014} design
the timing law as a function of $\theta$, the deviation ($\vx_d$ for
state feedback systems), and their higher derivatives:
\[
g(\theta^{(k)},\vx_d^{(k)},\ldots,\theta,\vx_d,u_0) = 0\,,
\]
wherein $\theta^{(k)}$ is the $k^{th}$ derivative of
$\theta$. However, defining the function $g$ is a nontrivial problem.

\section{PATH-FOLLOWING USING CLF}~\label{sec:main}
We now present the design of a path following scheme by specifying a
timing law as well as control for deviation from the reference
trajectory based on a control Lyapunov function.
Let us define a new coordinate system, in which the state of the
system is $\vz^t : [\theta, \vx_d^t]$, wherein
$\vx(t) = \vx_d(t) + \vx_r(\theta(t))$ is the original state of the
system. 
Also, the control inputs are $\vv^t : [u_0, \vu^t]$.  
We assume
$\theta$ is directly controllable using $u_0$: $\dot{\theta} = u_0$.
Therefore,
$\dot{\vx}_r = \frac{d \vx_r}{d\theta} \dot{\theta} = \vr(\theta) u_0$.
and thus
\[
\dot{\vx}_d = f(\vx_d + \vx_r(\theta), \vu) - \vr(\theta) u_0 \,.
\]

First, our goal is to design a control that seeks to stabilize
$\vx_d = 0$.  We define a CLF as a function $V$ over $\vx_d$, but independent
of $\theta$, that respects the following constraints:
\begin{equation}\label{eq:pf-clf}
\begin{array}{l}
	\noindent\textbf{(A)}:\ V (\vzero) = 0\ \mbox{and}\ (\forall\vx_d \neq \vzero)\; V(\vx_d) > 0 \\
	\noindent\textbf{(B)}:\ (\forall\ \theta,\ \vx_d \neq \vzero)\; (\exists \vu, u_0) \; \dot{V}(\theta, \vx, u_0, \vu) < 0\,.\\
\end{array}
\end{equation}
Note that although $V$ is a function of $\vx_d$, its derivative is a function of $\vx_d, \theta, u_0, \vu$:
\begin{align*}
\dot{V}(\theta, \vx_d, u_0, \vu) & = \nabla V(\vx_d) \cdot \dot{\vx}_d \\
 & = \nabla V(\vx_d) \cdot (f(\vx_d + \vx_r(\theta), \vu) - \vr(\theta) u_0).
\end{align*}

Conditions (A) and (B) are identical to those in
Definition~\ref{def:clf}, requiring the function $V$ to be positive
definite, and the ability to choose controls to decrease
$V$. Furthermore, note that the $(\forall\ \theta)$ quantifier in (B)
guarantees that this decrease is achieved no matter where the current
reference state $\vx(\theta)$ lies relative to the current state
$\vx_d$. Also, we note that the value of $u_0$ (rate of change for
$\theta$) is obtained through the feedback law derived from the CLF
$V$.

\begin{theorem}
  If there exists a function $V$ which respects Eq.~\eqref{eq:pf-clf},
  there exist \hadi{a feedback law and a timing law (Sontag formula~\cite{sontag1989universal})} that are smooth almost everywhere,
  and stabilize to the reference trajectory.
\end{theorem}
\iftrversion
\begin{proof}
  We appeal directly to Sontag's result to obtain an almost-everywhere
  smooth feedback $\vu(\vx_d, \theta)$ and $u_0(\vx_d, \theta)$ that
  guarantees that $\dot{V} < 0$ for all $(\vx_d, \theta)$ with
  $\vx_d \not= 0$~\cite{sontag1989universal}.

  Since $\dot{V} \leq 0$, and $V$ is radially unbounded, La Salle's
  theorem guarantees that the dynamical system will stabilize to the 0
  level set
  $V^{=0}:\ \{(\vx_d, \theta) |\ V(\vx_d) = 0 \} = \{ (0, \theta)\ |\
  \theta \in \reals \}$. Note however that $V^{=0}$ is a subset of the
  reference trajectory since $\vx_d = 0$.  
\end{proof}
\else
All proofs are available in~\cite{pathfollowing2018arxiv}.
\fi
Thus far, we have only demonstrated how CLFs can be used to decide on
a timing law as well as a control to nullify the deviation to
$\vzero$. However, this does not address a key requirement of progress: we
need to ensure that $\dot{\theta} > 0$ so that we make progress along
the reference from one end to another. Other limitations include that
saturation limits on
$\vu$ are not enforced, and that the avoidance of obstacles is not
considered. Finally, the stability is required to be global in the
entire state-space, which is very restricting.
All of these restrictions will be removed in the next section.

\section{PATH SEGMENT FOLLOWING PROBLEM}~\label{sec:bounded}
In the previous section, we discussed path following in a global
sense using CLFs, but noted several limitations. We will refine our
approach to address them in this section.

\paragraph{Finite Trajectories:} Consider a reference trajectory
\emph{segment} defined by $\vx_r(\theta) $ for
$\theta \in \Theta : [0, T]$. However, defining stability for such
segments is cumbersome. Therefore, we discuss reachability in finite
time along with safety (reach-while-stay).

We augment the given reference trajectory by adding an
initial set $\hat{I} \ni \vx_r(0)$, a goal set $\hat{G} \ni \vx_r(T)$,
and a parameterized family of safe sets
$\hat{S}(\theta) \ni \vx_r(\theta)$ for $\theta \in \Theta$,
such that $\hat{S}(0) \supseteq \hat{I}$ and
  $\hat{S}(T) \supseteq \hat{G}$.  Let
$\hat{S}:\ \bigcup_{\theta \in \Theta} \hat{S}(\theta)$ denote the
entirety of the safe set.  For convenience, we have defined
$\hat{S}, \hat{I}, \hat{G}$ in the original state space $\vx$. We will
now define them in terms of the deviation $\vx_d$ to define the
following sets:
\begin{align*}
I &: \{\vx_d \ | \ \vx_d+\vx_r(0) \in \hat{I}\} \\
G &: \{\vx_d \ | \ \vx_d+\vx_r(T) \in \hat{G}\} \\
S(\theta) &: \{\vx_d \ | \ \vx_d + \vx_r(\theta) \in \hat{S}(\theta) \}.
\end{align*}
Finally, let us denote $S:\ \bigcup_{\theta \in \Theta} S(\theta)$.

\paragraph{Input saturation:} 
Unlike infinite trajectories, here we assume initially $\theta = 0$
and in the end $\theta = T$. In this setting, we must ensure
progress for $\theta$ at each time-step. In other words, we wish to
make sure $\dot{\theta} > 0$. Recall that $\theta$ is controlled by a
virtual input $u_0$ ($\dot{\theta} = u_0$). Therefore, we need input
saturation for $u_0$ and we assume
$u_0 \in [\underline{u_0}, \overline{u_0}]$, where
$\underline{u_0} > 0$ and $\overline{u_0} < \infty$.  We also assume
the reference trajectory is feasible with respect to the dynamics. To
ensure that the reference trajectory remains feasible for the path
following problem, we enforce that 
$\underline{u_0} \leq 1 \leq \overline{u_0}$ as the timing law is simply
$\dot{\theta} = 1$ for the original reference trajectory
($\theta(t) = t$). Besides $u_0$, we will add saturation limits to the
inputs in $\vu$, as well. Formally we restrict $\vv \in \V$ for a
polytope $\V$.

\begin{definition}[Path Segment Following Problem]
  The path segment following problem, given
    $(S(\theta), I, G)$ and saturation constraints $\V$, is to derive
    a control feedback law $\vu = \vu(\theta, \vx_d)$ and timing law
    $u_0 = u_0(\theta, \vx_d)$ such that for all initial states
    $\vx_d(0) \in I, \theta(0) = 0$, the resulting state-control
    trajectory of the closed loop
    $(\theta(t), \vx_d(t), u_0(t), \vu(t))$ satisfies the following
    conditions: (a) saturation constraints are satisfied,
    $(u_0(t), \vu(t)) \in \V$ for all times $t$, (b) there exists $t^* > 0$
    such that $\vx_d(t^*) \in G$,
    i.e, the goal is reached, and (c) for all
    $t \in [0, t^*),\ \vx_d(t) \in S(\theta(t))$, while staying inside
    the safe set for all times until the goal is reached. Finally,
    note that condition (a) guarantees progress is made towards
    $\theta = T$ starting from $\theta = 0$.
\end{definition}

\paragraph{Proof Rules:} The control Lyapunov function argument can
get extended to control funnels for formally satisfying the
reach-while-stay property~\cite{bouyer2017timed}.
We will define a \emph{control funnel} as the sublevel sets of a smooth
function $V(\theta, \vx_d)$.
For a smooth function $V$, and a relational operator $\bowtie \in \{ <, \leq, =, \geq , >\}$,
let us define the following families of sets that are parameterized by $\theta$:
$V^{\bowtie \beta}(\theta) : \{\vx \ | \ V(\theta, \vx) \bowtie \beta\}$.
Furthermore, let $V^{\bowtie \beta} : \cup_{\theta \in \Theta} \ V_\theta^{\bowtie\beta} $.

\begin{definition}[Control Funnel Function]\label{def:control-funnel-function}
  A smooth  function
  $V(\theta, \vx_d)$ is called a control funnel function iff the
  following conditions hold:
\begin{equation}\label{eq:rules}
\begin{array}{lrl}
\mathbf{(a)} & (\forall \vx_d \in I) & \hspace{-0.2cm} V(0,\vx_d) < \beta \\
\mathbf{(b)} & (\forall \vx_d \not\in int(G)) & \hspace{-0.2cm} V(T,\vx_d) > \beta \\
\mathbf{(c)} & (\forall \theta \in \Theta, \vx_d \not\in int(S(\theta))) & \hspace{-0.2cm} V(\theta, \vx_d) > \beta \\
\mathbf{(d)} & (\forall \theta \in \Theta, \vx_d \in S(\theta) \cap V_\theta^{=\beta}) &  \\
& (\exists \vv \in \V) & \hspace{-0.2cm} \dot{V}(\theta,\vx_d, \vv) < 0.
\end{array}
\end{equation}
\end{definition}

\begin{figure}[t]
\begin{center}
\vspace{0.2cm}
	\includegraphics[width=0.4\textwidth]{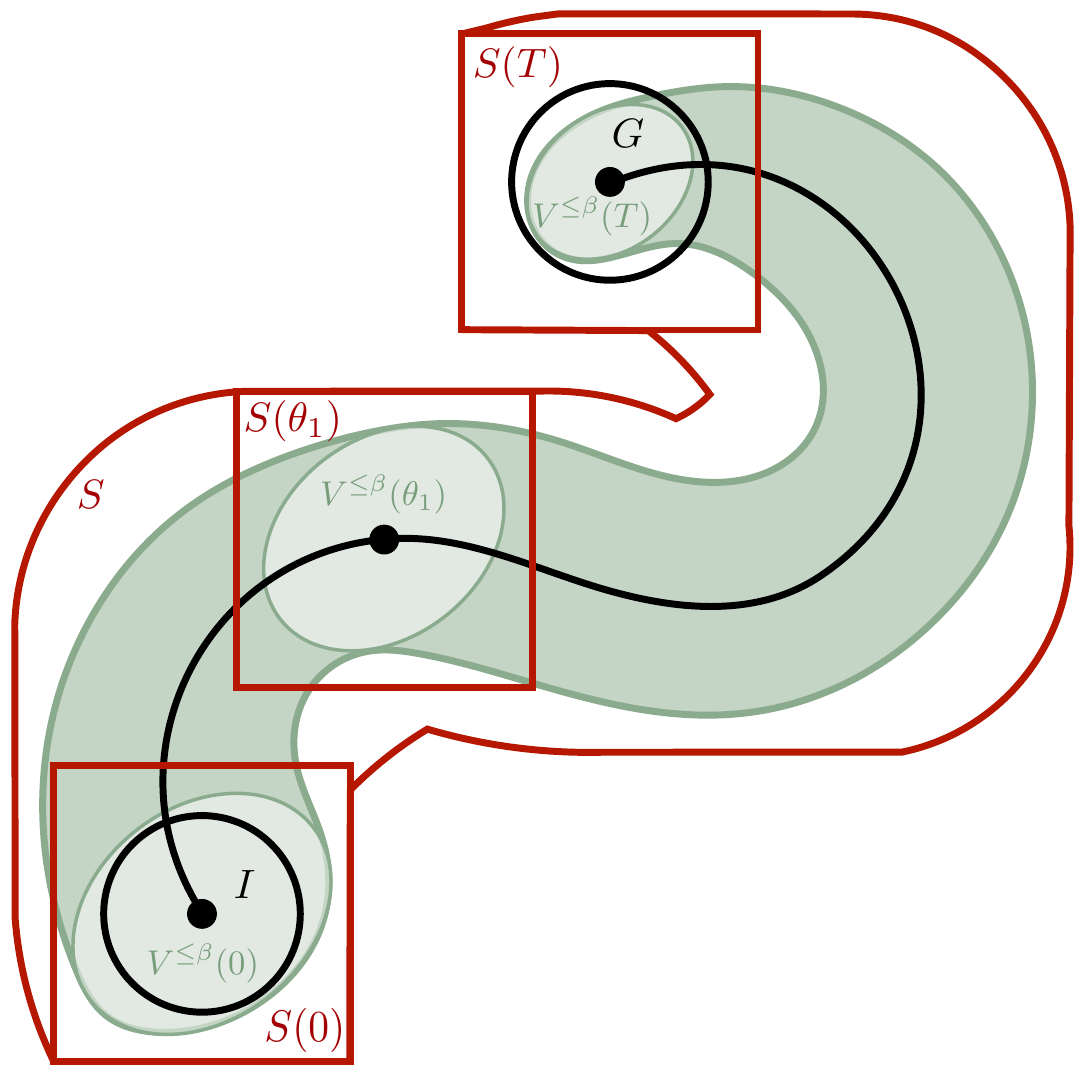}
\end{center}
\caption{Schematic View of a Funnel for reach while stay. $S(\theta_1)$ defines the safe set when $\theta = \theta_1$. $V(\theta) \leq \beta$ is an invariant.}\label{fig:funnel} 
\end{figure}

The idea, depicted in Fig.~\ref{fig:funnel}, is as follows. Initially (condition(a) in Eq.~\eqref{eq:rules}), $V < \beta$ ($\vx \in V^{<\beta}$). Condition (d) guarantees that for all the states in a neighborhood of set $V^{=\beta}$, there exist a feedback which decreases the value of $V$. Therefore, by providing a proper feedback, the state never reaches boundary of $V^{\leq \beta}$ ($V^{=\beta}$) because the value of $V$ can be decreased just before reaching $V^{=\beta}$. As a result, $V$ remains $< \beta$. This means the state stays inside $V^{< \beta}$ as long as $\theta \in \Theta$. Also the state remains in $S(\theta)$ as value of $V$ for other states is $\geq \beta$ (condition (c)). Since $\theta$ is increasing at minimum rate $\underline{u_0}$ at some point $\theta$ reaches $T$. Then, according to condition (b), the state must be in the interior of $G$ (top green ellipse), because otherwise value of $V$ would be $\geq \beta$.

\begin{theorem}
  Given a control funnel function $V$, there exist \hadi{a smooth feedback law and a timing law (Sontag formula~\cite{WIELAND2007462})}
  for reaching $G$ such that for any initial state $\vx_d(0) \in I$,
  the goal state is eventually reached at some time $t^*$ satisfying
  $T/\underline{u_0} \geq t^* \geq T/\overline{u_0}$, while staying in
  set $S$ for $0 \leq t \leq t^*$.
\end{theorem}
\iftrversion
\begin{proof}
	Initially $\vz(0) = [\theta(0), \vx_d(0)^t]^t$. Let $V(t) = V(\vz(t))$. According to condition (a), $V(0) = \beta_0 < \beta$ as $\theta(0) = 0$
    According to condition (d) in Eq.~\eqref{eq:rules} and Sontag formula~\cite{sontag1989universal,WIELAND2007462}, there exists a smooth feedback which decreases value of $V$ for all time instances that $\theta(t) \in \Theta \land \vx(t) \in S \land V(t) = \beta$. Also, by compactness of $V^{=\beta} \cap S$ it is guaranteed that under these circumstances, $\dot{V}(t) \leq -\epsilon$ for some $\epsilon > 0$. Moreover, there is a $\beta_1$ ($\beta_0 < \beta_1 < \beta$ s.t. if $\theta(t) \in \Theta \land \vx(t) \in S \land V(t) = \beta_1$, then $\dot{V}(t) \leq -\frac{\epsilon}{2}$.
	Now, we assume $\vx(\cdot)$ reaches boundary of $S$ before reaching $G$.
	Let $t_2$ be the first time instance that $\vx(\cdot)$ reaches the boundary of $S$. According to condition (c), $V(t_2) > \beta$. By smoothness of $V$ and the dynamics, there is a time $t$ ($< t_2$) for which $V(t) = \beta_1$. Let $t_1$ be the first time instance that $V(t_1) = \beta_1$ and $V^+(t_1) > \beta_1$. However, the feedback law forces $V$ to decrease at minimum rate $\frac{\epsilon}{2}$ which is a contradiction ($V^+(t_1) < \beta_1$). Therefore, either $\vx_d(\cdot)$ remains inside $R = V^{< \beta} \cap S$ forever or remains inside $R$ until it reaches $G$. On the other hand, let $t_f$ be the time $\theta(t_f) = T$ and $\frac{T}{\overline{u_0}} \leq t_f \leq \frac{T}{\underline{u_0}}$. Since $\vx_d(\cdot)$ remains in $V^{< \beta}$, $V(\vx_d(t_f)) < \beta$. According to condition (b), $\vx_d(t_f)$ is in the interior of $G$ and therefore $\vx(t_f) \in int(G)$.
\end{proof}
\fi

In practice, we replace $V^{=\beta}$ in condition (d) of Eq.~\eqref{eq:rules} with $V^{\geq\underline{\beta}}$ for some $\underline{\beta} \leq \beta$. Using this trick we make sure the value of $V$ can be decreased in a larger region to improve robustness. Also, any set $V^{\leq\hat{\beta}}$ (for $\hat{\beta} \geq \underline{\beta}$) would be an invariant until $\theta$ reaches $T$.

\paragraph{Increasing Robustness}
To improve robustness, one could simply maximize $\lambda$ \hadi{while feasibility is checked (Eq.~\eqref{eq:rules}) as the following:}
\begin{align*}
	Eq.~\eqref{eq:rules} \land & (\forall \theta \in \Theta, \vx_d \in S(\theta) \cap V^{=\beta}) \\ & (\exists \vv \in \V) \ \dot{V}(\theta,\vx_d, u_0, \vu) < -\lambda\,.
\end{align*}
The bigger is the $\lambda$, the faster the value of $V$ can be decreased, and therefore the resulting control law would be more robust.


%
%

\section{EXPERIMENTS}~\label{sec:expr}
In this section, we discuss the process of designing control funnels
for a bicycle model, followed by implementation and discussions.

\paragraph{Synthesizing Funnels:} We adapted a recently developed
demonstrator-based learning framework to synthesize control funnel
functions for given sets $I$, $G$, and
$S$~\cite{Ravanbakhsh-RSS-17}. \hadi{We note that it is possible to use SOS programming to design feedback and funnel function~\cite{majumdar2013control} to address the control design problem. However, SOS programming yields a bilinear matrix inequality, which comes with a lots of numerical issues and slow convergence, and seems to perform poorer. For a detailed comparison see~\cite{ravanbakhsh2018Learning}. 
}
Following that approach, the control
funnel function $V$ is parameterized as a linear combination of some
basis functions $V(\vz) : \sum_{j=1}^r c_j g_j(\vz)$. Next, we provide
an MPC-based demonstrator that given a concrete state
$(\theta, \vx_d)$, demonstrates an optimal control input $(u_0, \vu)$
by minimizing a cost function (distance to reference trajectory) for a
given finite time horizon. Furthermore, the conditions in Eq.~\eqref{eq:rules}
are checked for a given instantiation of parameters $(c_1, \ldots, c_r)$
using a verifier that uses a LMI-based relaxation.

We use the bicycle model presented in Eq.~\eqref{eq:bicycle-model}, but
use a ``body fixed frame'', wherein the state of the vehicle
is given by  $\vz^t : [\theta, \vx_R^t]$, and $\vx_R : [\alpha_R, x_R, y_R, v_R]^t$. The state variables
in the inertial frame $\vx(t) : [\alpha(t), x(t), y(t), v(t)]^t$ are written
in terms of $\vx_R$ as follows:
\[
\left[ \begin{array}{c}\alpha_R(t) + \alpha_r(\theta(t)) \\ \cos(\alpha_r(\theta(t))) x_R(t) - \sin(\alpha_r(\theta(t))) y_R(t) + x_r(\theta(t)) \\ \sin(\alpha_r(\theta(t))) x_R(t) + \cos(\alpha_r(\theta(t))) y_R(t) + y_r(\theta(t)) \\ v_R(t) + v_r(\theta(t)) \end{array} \right]\,.
\]
In this frame, $y_R$ axis is always aligned to axis of the vehicle in the reference trajectory.



%
%

We observe that the change of coordinates allows for accurate low
order polynomial approximations. Also, our experimental results
suggest that the learning framework succeeds in finding a control
funnel function of lower degree over the new coordinates when compared
to the inertial frame.  For all the experiments, we use the following
parameterization of $V$:
$V(\theta, \vx_R) : \vx_R^t C \vx_R\, + c_0 \theta$, where $c_0$ and
$C$ are the parameters to be synthesized.  For demonstration, we use
an off-the-shelf offline MPC with a simple quadratic cost
function. Please refer to~\cite{Ravanbakhsh-RSS-17} for more
details. As an alternative to Path-Following based Control Funnel
(PF-CF) we compare with Trajectory Tracking based Control Funnel
(TT-CF) obtained by setting $\dot{\theta} = 1$, and eliminating the
control input $u_0$.

\paragraph{Control Law Extraction:} For running the experiments, we need to extract control laws from control funnels. For a TT-CF, we merely use Sontag formula~\cite{sontag1989universal} with input saturation. Moreover, for a PF-CF, the controller stores and tracks value of $\theta$ (as a non-physical variable). The control law is extracted for both $\vu$ and $u_0$, and in addition to providing the feedback $\vu$, the controller updates the value of $\theta$ according to control input $u_0$.

\paragraph{Parkour Car}
\begin{figure}
\vspace{0.2cm}
\centering
	\includegraphics[width=0.7\columnwidth]{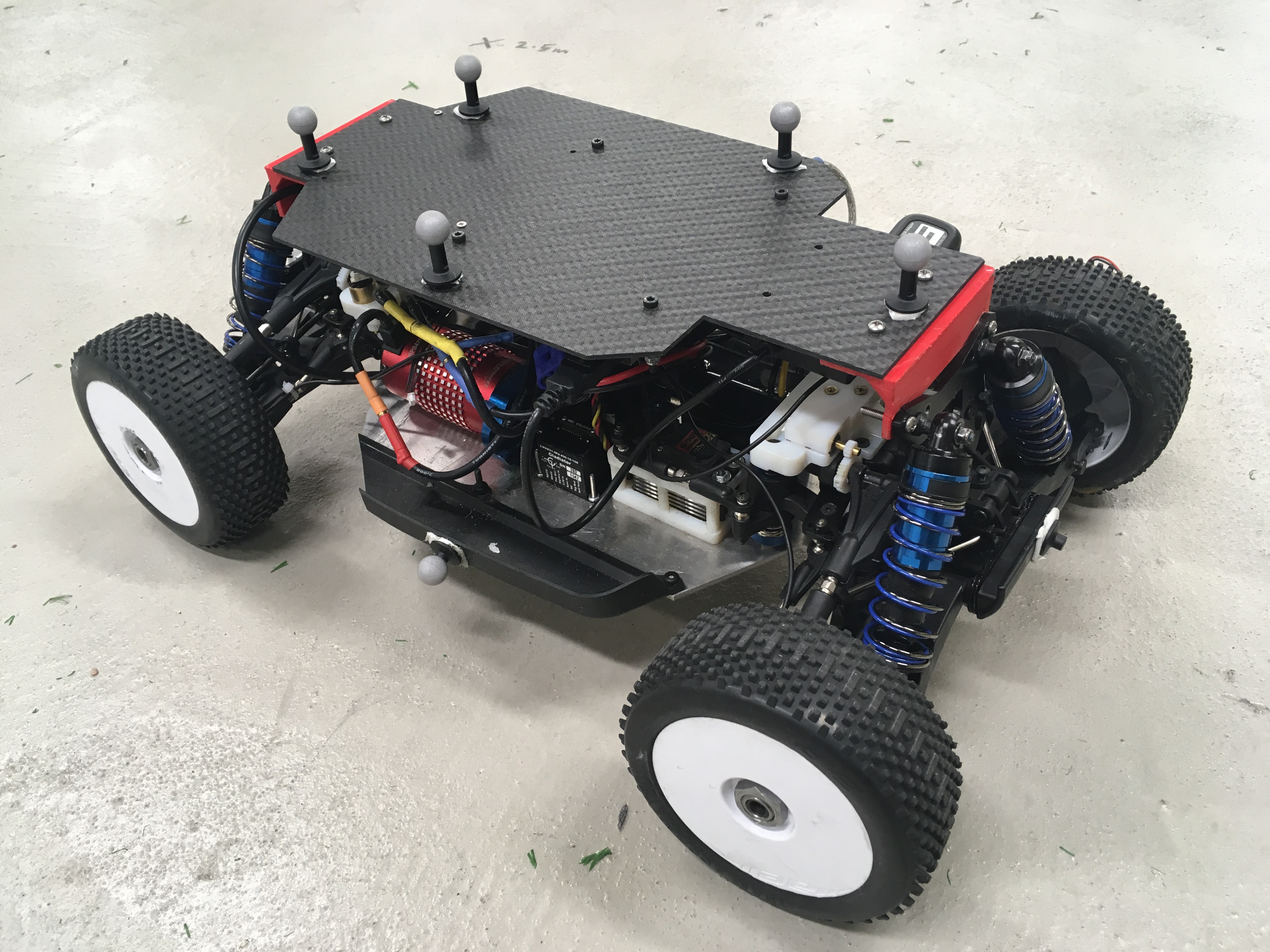}
\caption{Parkour car platform used for experiments.}\label{fig:parkourcar} 
\end{figure}
In order to verify functionality of proposed method we perform
experiments on a $\frac{1}{8}^{th}$ scale, four wheel drive vehicle
platform known as Parkour car (Fig.\ref{fig:parkourcar}) in a lab environment
equipped with a motion capture system. Parkour car has a wheel base of
$l = 34\text{cm}$ and includes an on-board computer to perform all computation on
the vehicle. While in action, the main computer receives a pose update
from motion capture system through WiFi connection, after which a new control
action is calculated based on the synthesized control law which then
gets transmitted to an ECU (Electronic Control Unit). The ECU handles
signal conditioning for acceleration and steering motors
on Parkour car. One iteration of this control action calculation can be
performed in less than $300\mu\text{s}$ on a single CPU core running at
3.5GHz. The low computation cost makes
this method attractive for real-time applications aboard platforms with low computation capabilities.

\paragraph{Straight Path:}
\begin{figure*}[t]
\begin{center}
\vspace{0.2cm}
	\includegraphics[width=1\textwidth]{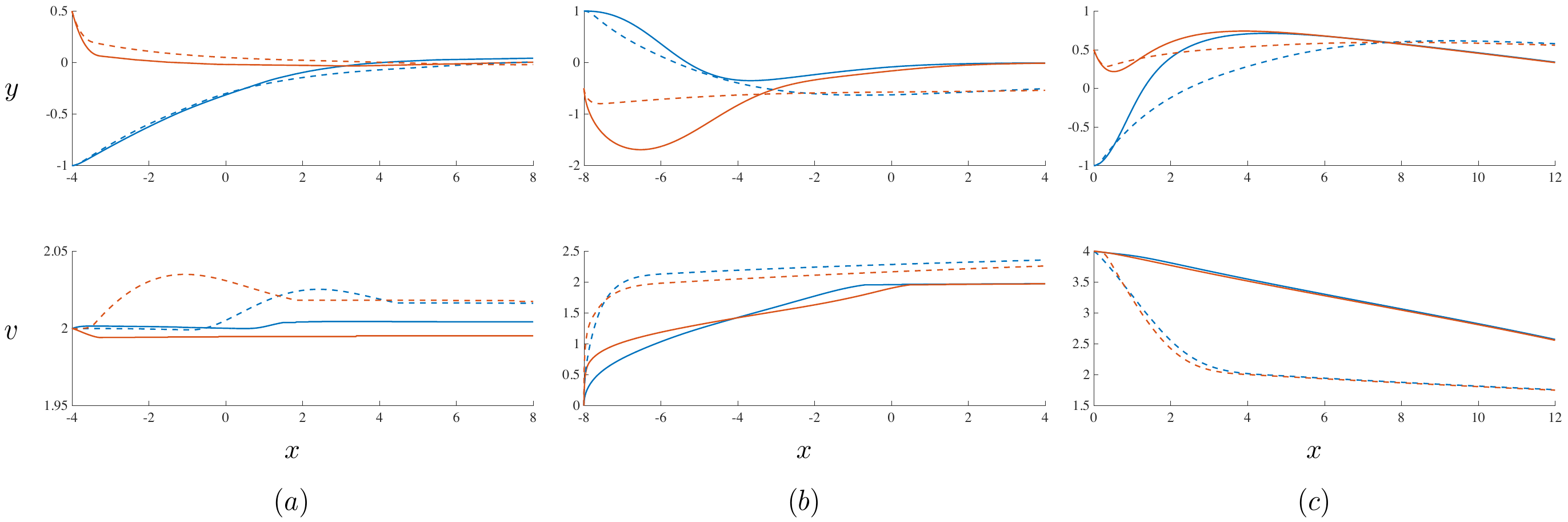}
\end{center}
Solid (dashed) lines are simulation trajectories corresponding to PF-CF (TT-CF). Blue (red) trajectories start from the same initial condition.
\caption{Simulation Results for a Straight Path}\label{fig:straight} 
\end{figure*}
In the first experiment, we consider a straight path from $x = -2$ to $x = 2$ with the reference trajectory $\vx_r(t) : [-\pi/2, -2+2t, 0, 2]^\intercal$. The sets are
\[ S(\theta) : [-1, 1]^3 \times [-3, 3] \, , \, I : \B_{0.5} \, , \, G
  : \B_{0.5} \,,\] where $\B_r$ is a ball of radius $r$ centered at the
origin.  The learning framework then successfully finds a
PF-CF. However, the learning framework fails to find a TT-CF. This
does not rule out the existence of a TT-CF, however.
Next, we increase the length of the path to $8\text{m}$ (from $x=-4$ to $x=4$). In this case, the learning framework can find a TT-CF. Fig.~\ref{fig:straight} shows simulation trajectories corresponding to the PF-CF and the TT-CF. For comparison, starting from the same initial conditions, the simulation is performed until $x$ reaches $x(0) + 12$. Fig.~\ref{fig:straight}(a) shows the results for initial states where the initial state is near $I$. The simulations suggest that both methods perform similarly and all trajectories converge to the path ($y$ converges to zero). The simulation time for all cases are similar and around $6\text{s}$. 
Also, the velocity of the vehicle is almost constant for both methods. Fig.~\ref{fig:straight}(b) shows the results for cases when the initial states are further away from $G$ (it needs more forces/time to reach $G$). In this case, the path-following method takes a longer time to reach $x = 4$ as the speed increases smoothly.
Fig.~\ref{fig:straight}(c) considers initial states that are closer to $G$. For these case, the path-following method takes a shorter time to reach $x = 12$ as the speed decreases smoothly.
The results demonstrate that the path-following method yields a faster convergence to the reference path. Moreover, the velocity changes smoothly while the trajectory tracking method settles the target velocity immediately.

We also investigate the same problem (straight path from $x = -4$ to $x = 4$) where the velocity is more restricted:
\begin{align*}
S(\theta): & [-1,1]^3 \times [-0.5, 0.5] \\
	I : G : & \{\vx | 4\alpha^2 + 4x^2 + 4y^2 + 16v^2 \leq 1\} \,.
\end{align*}
Again, under these circumstances, learning TT-CF fails while finding PF-CF is feasible. In other words, in trajectory tracking the change of velocity is crucial for reducing the tracking error.

\paragraph{Circular Path}
To carry out experiments on Parkour car and examine the behavior over long trajectories, we consider a circular path with radius $1.5\text{m}$. The vehicle moves with a constant velocity $\frac{\pi}{2} m/s$ and the reference trajectory would be $\vx_r(t) : [\frac{\pi}{3}t, 1.5 \cos(t), 1.5 \sin(t), \frac{\pi}{2}]^t$. For the learning process, we consider a finite trajectory (once around the circle) where $t \in [0, 6]$. The sets are:
\begin{align*}
    S(\theta) : [-1, 1] \times [-3, 3], \ I : \B_{0.5}, \ G : \B_{0.5}\,.
\end{align*}

\begin{figure}[t]
\begin{center}
    \includegraphics[width=0.45\textwidth]{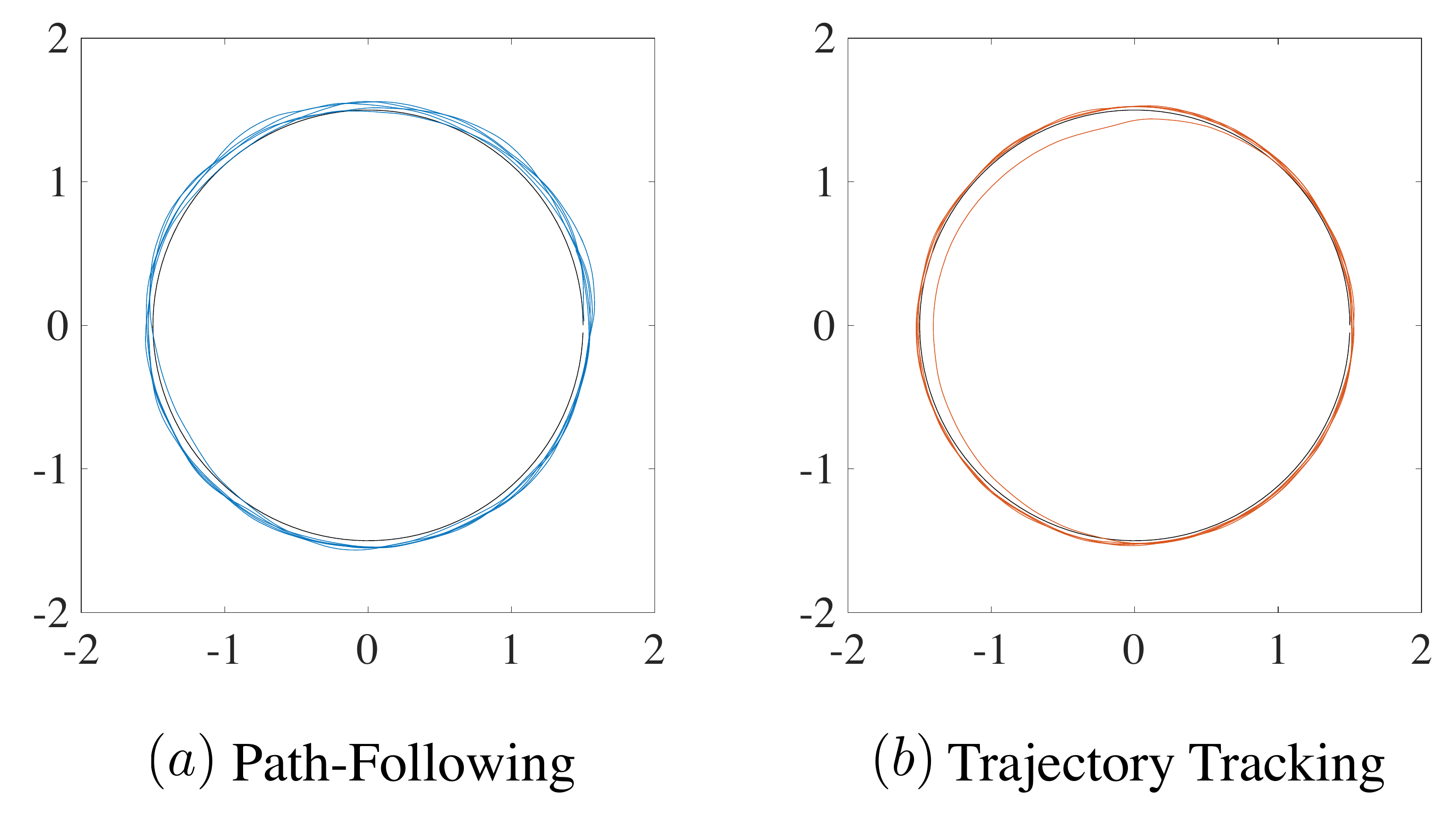}
\end{center}
\caption{Projection of trajectories, generated on Parkour car, for the circular path. Parkour car finishes five rounds around the circle. The reference trajectory is shown in black.}\label{fig:circular-5rounds} 
\end{figure}

\begin{figure}[t]
\begin{center}
    \includegraphics[width=0.45\textwidth]{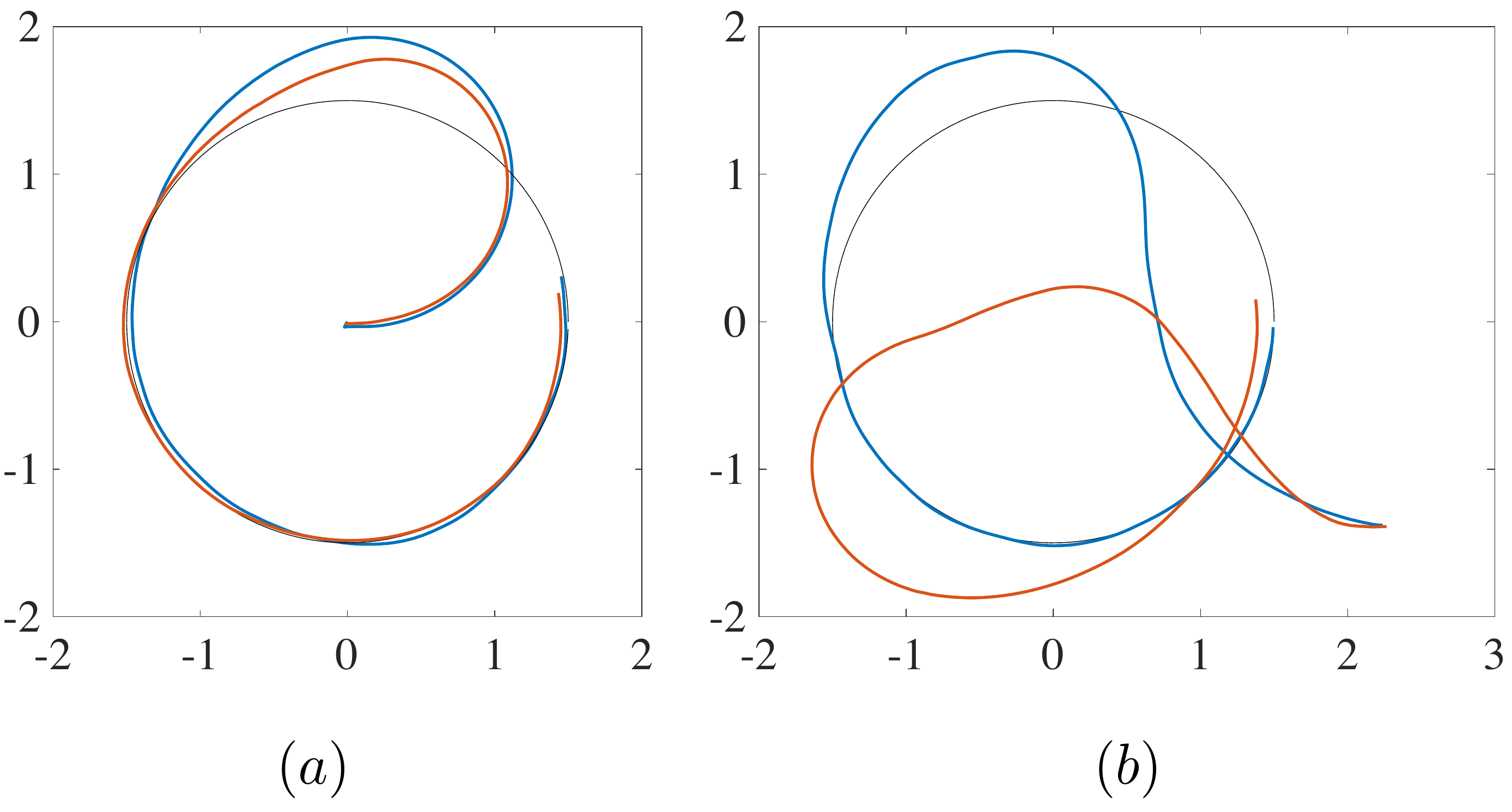}
\end{center}
\caption{projection of trajectories, generated on Parkour car, for the circular path form different initial states. Blue (red) lines corresponds to the path-following (trajectory tracking) method. The reference trajectory is shown in black. Initial state: (a) $[-\pi/2, 0, 0, 0]$ and (b) $[\pi, 2.25, -1.4, 0]$.}\label{fig:circular-inits} 
\end{figure}

Figure~\ref{fig:circular-5rounds} shows the trajectories when the controller runs on Parkour car. Despite the uncertainties in the measurements and simple modeling, both controllers do a good job of following the reference path. Fig.~\ref{fig:circular-inits} shows the trajectories for different initial states. Fig.~\ref{fig:circular-inits}(b) suggest that the trajectory tracking method may take shortcuts to satisfy time constraints.

We also investigated the same problem with higher reference velocity. When the reference velocity is increased to $\pi$ (from $\pi/2$), we could not find a TT-CF. Nevertheless, increasing reference velocity does not seem to affect the process of finding PF-CF, and we can discover solutions even if the reference velocity is $10\pi$.

\begin{figure}[t]
\begin{center}
\vspace{0.2cm}
    \includegraphics[width=0.23\textwidth]{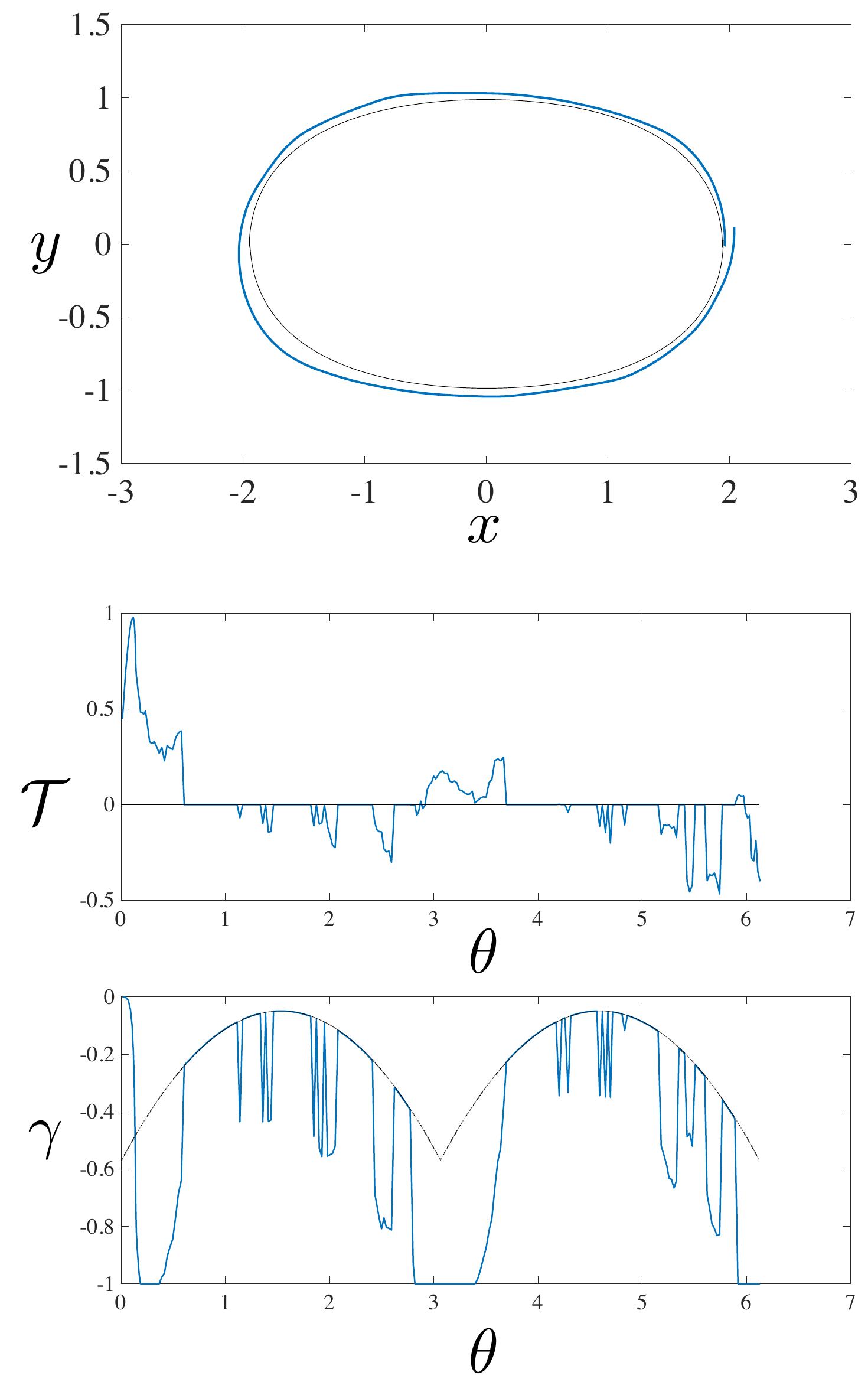}
\end{center}
\caption{Projection of the trajectory, generated on Parkour car, for the oval path. The reference trajectory is shown in black.}\label{fig:oval-inputs} 
\end{figure}

\paragraph{Oval Path:} Following a circular path is easy as the curvature remains fixed. However, the problem is more challenging when the path is an oval. The goal is to follow an oval path $P : \{\frac{y^2}{1^2} + \frac{x^2}{2^2} = 1\}$. First, a reference trajectory is generated to follow this path closely. As polynomial approximations of the reference path become more challenging, we divide the reference path into two similar parts. Then, we find a funnel for each part and make sure we can concatenate these two funnels. For the first part, the goal is to reach from $\B_{0.5}([2, 0])$ to $\B_{0.5}([-2, 0])$ going in a CCW direction and then reach from $\B_{0.5}([-2,0])$ to $\B_{0.5}([2, 0])$ again in a CCW direction. For both segments we use the following sets:
\begin{align*}
    S(\theta) : [-1, 1] \times [-3, 3], \ I : \B_{0.5}, \ G : \B_{0.5}\,.
\end{align*}
Notice that since $G$ for the first segments fits in $I$ for the second segment, we can safely concatenate the funnels. If a trajectory tracking method is being used, the learning procedure fails to find solutions. However, the path following method yields proper control funnels. Figure~\ref{fig:oval-inputs} shows the trajectories generated from our experiments using the CF-based controller. The tracking is not precise when the curvature is at its maximum. We believe the main reason is input saturation for the steering, which occurs because of the imprecise model we use (Fig.~\ref{fig:oval-inputs}).

\begin{figure}[t]
\begin{center}
\vspace{0.2cm}
	\includegraphics[width=0.3\textwidth]{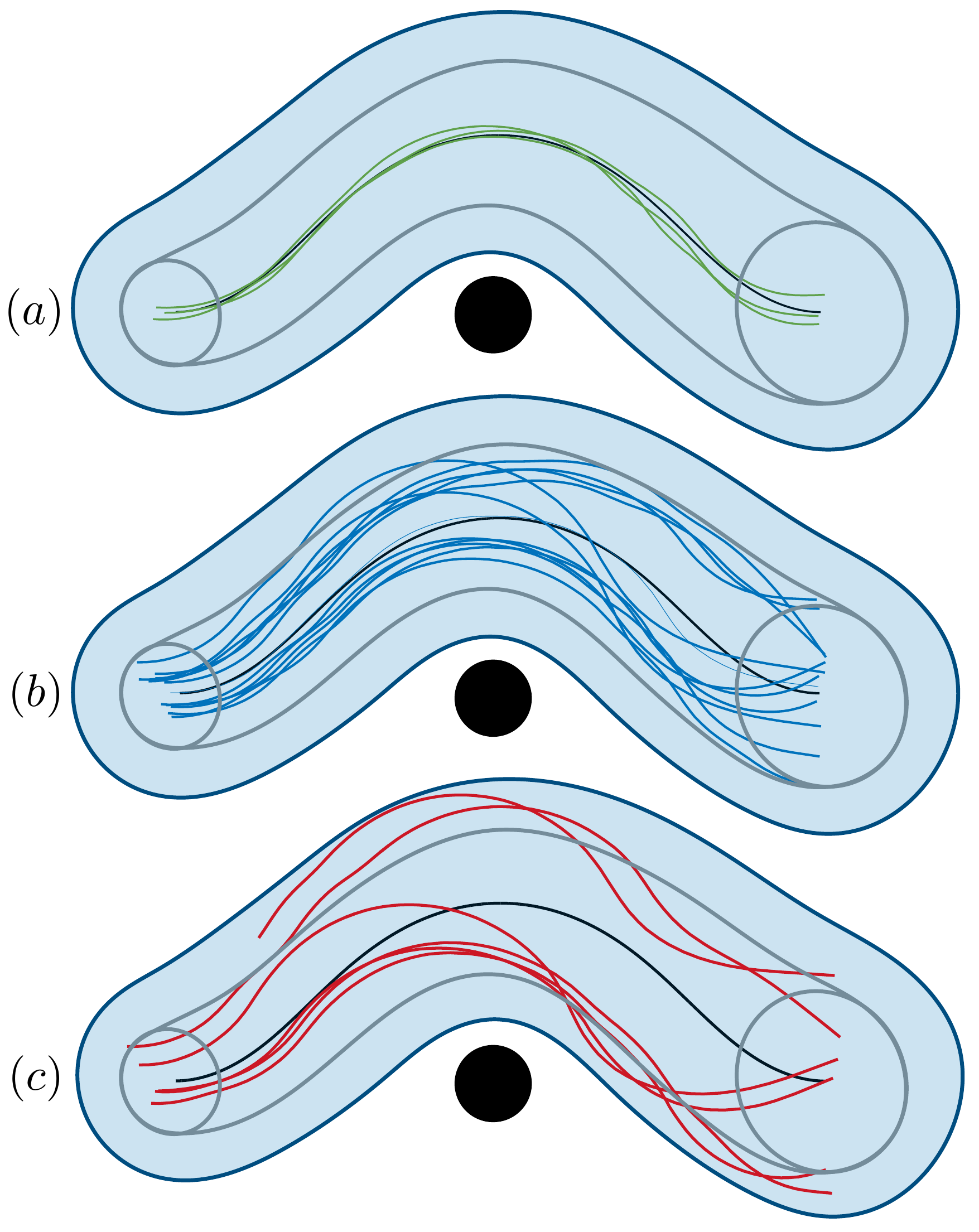}
\end{center}
\caption{Projection of trajectories on $x--y$ plane generated by Parkour car for the obstacle avoidance problem. Funnel boundary is shown in gray and as long as the center of car is in the funnel, the whole body of the car remains in the blue region. (a) guaranteed traces, (b) not guaranteed but safe traces, (c) not guaranteed and unsafe traces. }\label{fig:obs-trace} 
\end{figure}

\paragraph{Obstacle Avoidance:} Going back to the scenario of obstacle avoidance, we wish to find a control funnel to guarantee safety (avoiding the obstacle). Having a reference trajectory, instead of defining $S(\theta)$, we simply define $S$ as
\[
S:\ \{ \vx \ | \ ([x, y] \oplus \B_{0.25}) \cap O = \emptyset \}\,,
\]
where $O$ is the obstacle, $\oplus$ is the Minkowski sum, and $0.25m$
is the distance between the center of the car and its corners (the
body of Parkour car fits in $\B_{0.25}$). This trick allows to
reason only about the center of the car, and safety is guaranteed as
long as the center of the car is in $S$. Next, we set $I : \B_{0.25}$
and $G : \B_{0.25}$. However, we can not find a solution using the
learning framework. To relax the conditions, we allow $G$ to be larger
$G :\B_{0.5}$. In this case, we were able to find a solution (only if
the path-following method is being used). For the experiment,
Parkour car moves toward the obstacle with different initial states
and the CF-based controller engages when the car is $1.5m$ to the left
or right of the obstacle's $x$ position. Fig.~\ref{fig:obs-trace}
shows the projection of the funnel on $x$-$y$ plain. We note that if a
trajectory starts from the head of the funnel, not only its initial
$x$ and $y$, but also its initial $v$ and $\alpha$ should also be
inside the funnel. Fig.~\ref{fig:obs-trace} (a) shows trajectories
where the initial state is inside the head of the funnel. As shown,
the trajectories remain inside the funnel and reach the tail. However,
as demonstrated in Fig.~\ref{fig:obs-trace} (b), even if the
trajectory starts outside of the funnel head, the whole body of the
car may remain in the guaranteed region (blue region). Nevertheless,
the safety is not guaranteed any longer as Fig.~\ref{fig:obs-trace}
(c) shows trajectories where Parkour car leaves the guaranteed
region.

\section{CONCLUSIONS}~\label{sec:conclusion}
In this work, we investigate the use of control Lyapunov functions for
path following and provide a characterization of control funnel
functions for tracking a trajectory segment. Our approach lends itself
to an efficient synthesis technique presented previously. We implement
the resulting controller on the Parkour car and show its effectiveness
through a set of tracking problems. Future work will focus on integrating
our approach more closely with planning approaches to augment existing
approaches to the synthesis of control funnels~\cite{majumdar2013robust}.

\section*{ACKNOWLEDGMENT}
This work was funded in part by NSF under award numbers SHF 1527075 and CPS 1646556. All opinions expressed
are those of the authors and not necessarily of the NSF.


\bibliographystyle{IEEEtran}
\bibliography{refs}

\end{document}